\documentclass[letterpaper]{article} 
\usepackage[submission]{aaai23}  
\usepackage{times}  
\usepackage{helvet}  
\usepackage{courier}  
\usepackage[hyphens]{url}  
\usepackage{graphicx} 
\usepackage{natbib}  
\usepackage{caption} 
\usepackage{algorithm}
\usepackage{algorithmic}

\usepackage[pagebackref=true,breaklinks=true,colorlinks=true,bookmarks=false,linkcolor={red!50!black},urlcolor={darkgray!80!black},citecolor={green!50!black}]{hyperref} 
\usepackage{url}            
\usepackage{booktabs}       
\usepackage{amsmath}
\usepackage{nicefrac}       
\usepackage{microtype}      
\usepackage{xcolor}         
\usepackage{graphicx}
\usepackage{tabularx}
\RequirePackage{enumitem}
\usepackage{amsfonts}       

\usepackage{newfloat}
\usepackage{listings}
\DeclareCaptionStyle{ruled}{labelfont=normalfont,labelsep=colon,strut=off} 
\floatstyle{ruled}
\newfloat{listing}{tb}{lst}{}
\floatname{listing}{Listing}
\newcommand{\Tref}[1]{Table~\ref{#1}}

\newcommand{\fref}[1]{Fig.~\ref{#1}}
\newcommand{\Fref}[1]{Figure~\ref{#1}}

\newcommand{\eg}{\textit{e}.\textit{g}.}
\newcommand{\ie}{\textit{i}.\textit{e}.}

\title{PartNerFace: Part-based Neural Radiance Fields for Animatable Facial Avatar Reconstruction}
\author {
	Xianggang Yu, 
	Lingteng Qiu, 
	Xiaohang Ren, 
	Guanying Chen, \\
	Shuguang Cui, 
	Xiaoguang Han\textsuperscript{$\dagger$}, 
	Baoyuan Wang
}

\begin{document}

\twocolumn[{%
	\renewcommand\twocolumn[1][]{#1}%
	\maketitle
	\vspace{-14mm}
	\begin{center}
		\centering
		\captionsetup{type=figure}
		\includegraphics[width=.9\textwidth]{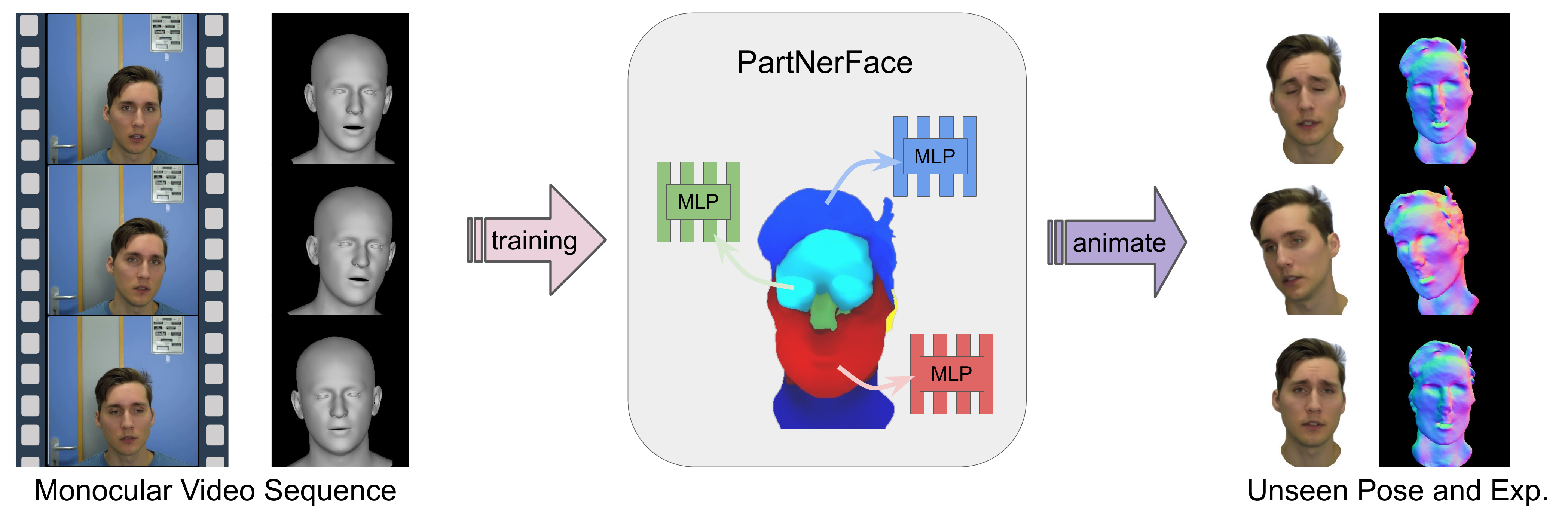}
		\captionof{figure}{Given a monocular human face video, the proposed \textit{PartNerFace} reconstructs a facial avatar by learning a part-based neural radiance fields conditioned on the estimated FLAME coefficients, which can be animated with novel poses and expressions.}
		\label{fig:teaser}
	\end{center}%
}]



\begin{abstract}
	We present PartNerFace, a part-based neural radiance fields approach, for reconstructing animatable facial avatar from monocular RGB videos.
	Existing solutions either simply condition the implicit network with the morphable model parameters or learn an imaginary canonical radiance field, making them fail to generalize to unseen facial expressions and capture fine-scale motion details.
	To address these challenges, we first apply inverse skinning based on a parametric head model to map an observed point to the canonical space, and then model fine-scale motions with a part-based deformation field.
	Our key insight is that the deformation of different facial parts should be modeled differently. 
	Specifically, our part-based deformation field consists of multiple local MLPs to adaptively partition the canonical space into different parts, where the deformation of a 3D point is computed by aggregating the prediction of all local MLPs by a soft-weighting mechanism.
	Extensive experiments demonstrate that our method generalizes well to unseen expressions and is capable of modeling fine-scale facial motions, outperforming state-of-the-art methods both quantitatively and qualitatively.
	
\end{abstract}

\section{Introduction}

Reconstructing animatable facial avatars from monocular RGB videos is an important yet highly challenging and ill-posed problem. It requires synthesizing high-fidelity face appearance under various viewpoints, poses, and expressions, which has been the foundation of many promising applications such as facial VR/AR, teleconference, games and movie production. 

Currently, a plethora of approaches have been proposed to tackle this problem. Traditional solutions often resort to model-based approaches that optimize the per-frame 3D morphable model (3DMM)~\cite{paysan20093d} parameters through a dense analysis-by-synthesis process~\cite{thies2016face2face, Cao2015RealtimeHF, Cao20133DSR, Garrido2016ReconstructionOP, Deng2019Accurate3F}.  However, the statistic models they employed can only express the geometric variation of the face region, while some non-facial accessories, such as hair and glasses, are hard to be modeled.
Another line of work proposes 2D motion-based methods by learning a relative motion field from the source to driving and utilize generative network for face image synthesis~\cite{siarohin2019animating, wang2021safa,Zakharov2019FewShotAL,Wang2021OneShotFN}. Compared with 3DMM-based methods, 2D motion-based methods can better generalize to
handle the synthesis of hair, beard, and other accessories. Nevertheless,
their results usually suffer from `flat' appearance since 2D motion filed is itself limited to model out-of-plane transformations 
~\cite{wang2021safa}.


Recently, witnessing the success of neural radiance fields (NeRF)~\cite{mildenhall2020_nerf_eccv20}, several state-of-the-art methods extend the static NeRF method to model dynamic facial avatars. The key challenge for this extension is how to precisely model the pose changes and non-rigid motions of human face using implicit neural representations. Prior works either condition the deformation onto a per-frame pre-fitted 3DMM parameters~\cite{gafni2021dynamic} (more specifically, camera pose and expression basis); or predict the deformation field between each observation and an imaginary canonical NeRF model that shares among all observations~\cite{park2021_nerfies_iccv21}. However, due to the insufficient poses and expressions presented in the training corpus, the parameter-conditioning method is prone to overfitting and suffers from the ``pose-expression'' correlation problem, \ie, the network memorizes the specific pattern of ``pose-expression'' pairs in the training data. When facing with an unseen ``pose-expression'' pair during reenactment, the synthesized results are far from satisfactory. 
On the other hand, the latter method based on the canonical model has better generalization ability. Nevertheless, while faithfully reconstructing the input video, their imaginary canonical NeRF model exhibits less flexibility for control, and the trained model cannot be reenacted with novel poses and expressions. More importantly, none of the recent SOTA methods take into account the inherently part-based speciality of human face, and they all utilize a global generator function which is known to lead to representations that can not model fine-scale geometric details~\cite{egger20203d}.

In this paper, we propose \textit{PartNerFace}, a novel \textit{part-based} neural radiance fields method, for animatable facial avatar reconstruction from monocular RGB videos. 
Our key insight is that the deformation of different parts of human face exhibits less ``inter-region'' similarities but stronger ``intra-region'' correlations. For example, a smiling face will cause large deformation on the mouth and cheek regions, while the upper face like eyelids and forehead remains nearly motionless. Except facial features, other properties of a facial avatar such as hair and accessories are also related to different face parts, \eg, hair sticks to the scalp part while glasses are mainly coupled with the eyes. Based on these observations, we propose to learn a novel \textit{part-based deformation field} to model the deformation of different facial parts separately. 
Specifically, we learn a face space partition with an \textit{adaptive part assigner} to predict the probability of any 3D point's part ascription, which is analogous to a continuous semantic fields, and employ $N$ (empirically set to 7 in our experiments) local MLP-based deformation networks to predict a deformation of any 3D point belong to each part.
Afterwards, the output from $N$ local deformation networks are aggregated by the predicted probability in a soft-weighting mechanism. Note that despite optimizing fully unsupervised, the adaptive part assigner is able to produce reasonable face partitions.
On the other hand, following FLAME~\cite{li2017learning}, we model the pose and expression variations of human face in canonical space, and map each point in posed space into canonical space with a coarse-level skinning-based deformation and a fine-level part-based neural deformation, conditioning on the FLAME coefficients. This formulation naturally offers explicit control and circumvents the overfitting issue, as we learn the radiance fields in canonical space which can generalize to unseen pose and expressions.

To summarize, this work has the following contributions:
\begin{itemize}[leftmargin=*,topsep=0pt]
	\item We introduce a novel part-based neural radiance field for facial avatar reconstruction, which not only generalizes well to unseen poses and expressions, but also models fine-scale facial motions. The key to our method is the learning of a part-based deformation field, which has not been used in prior works for avatar modeling. 
	\item We propose a novel adaptive part assigner along with a staged training strategy that produce plausible facial partitions while trained in a weakly-supervised manner.
	
	\item Our approach demonstrates compelling reconstruction results and significant performance improvement on extrapolation to novel poses and expressions.
\end{itemize}

\section{Related Works}
%
%
%

\paragraph{Dynamic Facial Modeling}

Recent dynamic facial modeling approaches can be roughly divided into 3D-reconstruction based and 2D-GAN based approaches. 
3D-reconstruction based approaches \cite{thies2016face2face, geng20193d, thies2020neural} reconstruct the source face from videos using a morphable face model \cite{paysan20093d, booth20163d, li2017learning} and forward render the target facial motion by morphing its geometry. 
Benefit from the recent development of neural renderers, \cite{deng2019accurate, lin2020towards, zhu2020reda, feng2021learning} add texture reconstruction into morphable face model to obtain accurate face modeling from a single source face RGB image. 
2D-GAN based approaches \cite{wang2019fewshotvid2vid, qiao2018emotional, song2018geometry} modeling the source face using generative networks to reconstruct video face frames and output photo-realistic face animation by changing its 2D facial motion such as 2D facial landmarks \cite{qiao2018emotional, song2018geometry, zakharov2019few, wu2018reenactgan} and latent motions \cite{wiles2018x2face, ha2020marionette}. 
Based on the learnt warping field, approaches \cite{siarohin2019animating, zakharov2020fast, zhao2021sparse, wang2021safa} warp feature maps of the generator to animate the face from a single image. 
First Order Motion Model (FOMM) \cite{siarohin2019animating} learns a general image warping field generator for face modeling and deforms the source image by the target motion warping field. 
SAFA \cite{wang2021safa} combines 3D face reconstruction and 2D GAN by adding the dense motion information of parametric face model into 2D GAN and enhances motion reproducibility. Our paper also targets 4D facial modeling, we try to use the newly-invented Neural Radiance Field to address the challenges of existing approaches.

\begin{figure*}[!t] \centering
    \includegraphics[width=.95\textwidth]{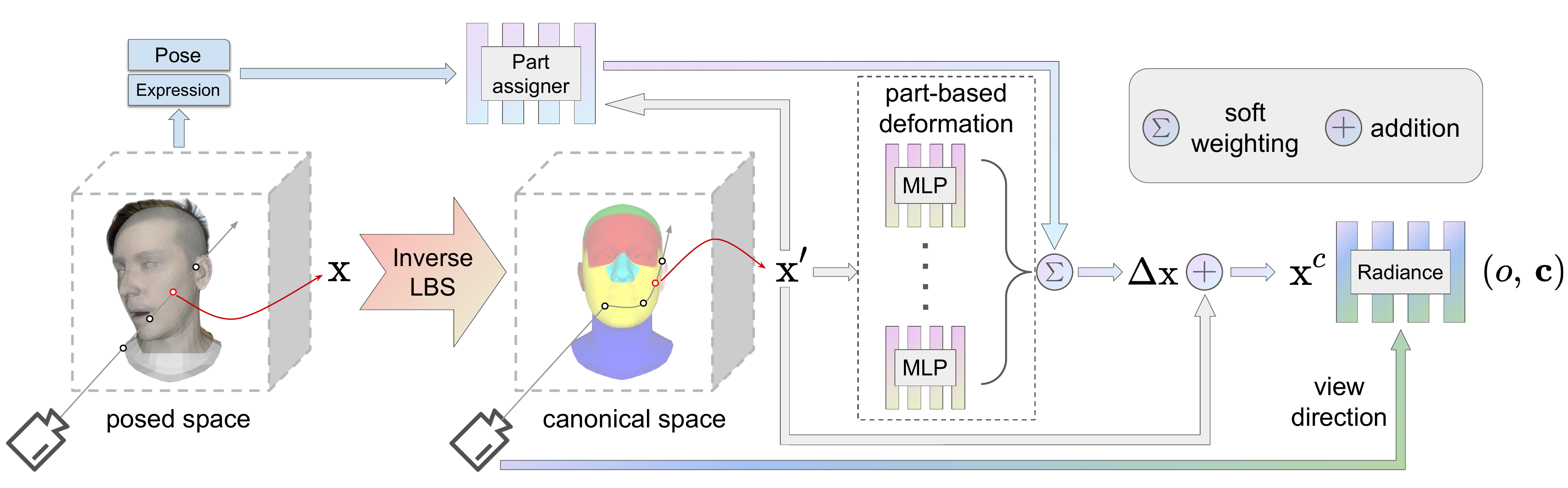}
    \caption{\textbf{Overview of PartNerFace}. Our method represents the facial avatar as a neural radiance fields in the canonical space. Given a 3D point in the posed space, it is first transformed to the canonical space through inverse LBS based on the FLAME model, then followed by a part assigner which predicts the probability of its part ascription. This probability is used to aggregate the output from a set of local deformation networks. The weighted deformation is then added to the previous transformed point. Finally, the occupancy and color is evaluated in the canonical space for volume rendering.} 
    
    \label{fig:pipeline}
    \vspace{-5mm}
\end{figure*}

\paragraph{Dynamic Neural Radiance Fields}
Inspired by the success of neural radiance fields (NeRF)~\cite{mildenhall2020_nerf_eccv20} in static scene modeling, many efforts have been devoted to extend NeRF to model dynamic scenes~\cite{pumarola2021_dnerf_cvpr21,tretschk2020_nonrigid_arxiv,li2021_nsff_cvpr21,xian2021_space_cvpr21}. 
A series of methods model the dynamic scene as a 4D spatial-temporal function~\cite{li2021_nsff_cvpr21,xian2021_space_cvpr21,du2021_nerflow_iccv21}.
Another branch of methods adopt a canonical space representation for shape and appearance modeling, and additionally learn a deformation field~\cite{pumarola2021_dnerf_cvpr21,tretschk2020_nonrigid_arxiv,park2021hypernerf} to model motions.
The canonical-based framework has been adopted by many dynamic human modeling methods~\cite{peng2021neural,peng2021animatable,liu2021neural,su2021nerf}, where the skeleton-based SMPL model~\cite{SMPL:2015} is used as a prior to model human motion.

In the problem of dynamic face modeling, Nerfies~\cite{park2021_nerfies_iccv21} learns an imaginary canonical space along with a deformation field to support dynamic video rendering, but it is not designed for animation.
NerFace~\cite{gafni2021dynamic} conditions the MLP network on the 3DMM parameters to model face motion and targets to facial animation driving. However, due to its inherent overfitting issue, it shows poor generalization capability to unseen poses and expressions.
Recently, several SOTA methods \cite{zheng2021avatar,grassal2021neural} also adopt FLAME model for dynamic face modeling. Neural head avatar~\cite{grassal2021neural} learns a pose-dependent geometry refinement network through explicit rendering and targets on reconstructing a detailed face mesh, while IMavatar~\cite{zheng2021avatar} uses forward skinning for global deformation modeling and requires a computationally expensive root-finding operation. 
Differently, our part-based deformation idea is orthogonal to these SOTA, and models the facial motion better and produces more photo-realistic renderings.

\paragraph{Part-based Facial Modeling}
Many real-life subjects are naturally part-based, and human face is one of the representatives. Since different facial parts usually exhibit different properties, globally modeling human face tends to produce reconstruction errors. Therefore, the part-based mechanism is introduced into the facial modeling literature. Some pioneer works~\cite{blanz1999morphable, decarlo2000optical, basso2008fitting} manually partition the face into several parts and model each part separately.  Their experimental results demonstrate that part-based modeling can produce higher fidelity geometry. The partition approaches are later on evolved to be automatically \cite{joshi2006learning, golovinskiy2006statistical, zhang2008spacetime, smet2010optimal, neumann2013sparse, ferrari2015dictionary}. As its advantage of local-based modeling, such a strategy is also adopted for facial motion capture \cite{wu2016anatomically}. 

\newcommand{\R}{\mathbb{R}}

\section{Methodology}












\paragraph{Overview}
Given a monocular RGB human face video, the proposed \emph{PartNerFace} aims at reconstructing an animatable facial avatar that can be rendered from novel pose and expressions. 
Specifically, Our method represents the shape and appearance of the face in a canonical radiance field, and transforms 3D points from each posed space into the canonical space in a coarse-to-fine manner for color rendering (see~Fig.~\ref{fig:pipeline}).
Specifically, we first apply inverse linear blend skinning (LBS) based on a parametric head model to model the coarse-level motions, and then adopt a part-based deformation field to capture the fine-scale motions of human face.
Volume rendering is performed on the transformed points in the canonical space to render image colors.

%

\subsection{Canonical Radiance Fields}
\label{subsec:canonical}

The static NeRF represents a scene as a continuous volumetric representation. Given a 3D location $x \in \mathbb{R}^3$ and a view direction $d \in \mathbb{R}^2$, the continuous representation maps them into the emitted color and density. To make NeRF modeling a dynamic face video, we assume a canonical radiance fields that shared across all frames. 
Specifically, we represent the canonical neural radiance fields with a multi-layer perceptron (MLP) $F_{c}$. Following~\cite{oechsle2021unisurf}, we parameterize $F_{c}$ to output color $\mathbf{c} \in \R^3$ and occupancy $o$ given a canonical point position $\mathbf{x^c}\in \R^3$:
\begin{align}
	(o, \mathbf{c}) &= F_{c}\left(\mathbf{x^{c}}\right)
	\label{func:canonical_model}
\end{align}
where $o \in [0, 1]$ indicates the probability that $\mathbf{x^c}$ is occupied in canonical space.
For each frame in the video, we predict the shape coefficients $\beta$, pose coefficients $\theta$, and expression coefficients $\psi$ of the FLAME model using EMOCA~\cite{danecek2022emoca}, and for any point $\mathbf{x}$ in the posed space, its corresponding canonical position $\mathbf{x^c}$ is computed as:
\begin{equation}
\begin{aligned}
	\mathbf{x^{c}} &= D_{ilbs}\left(\mathbf{x}, \beta, \theta, \psi\right) + \\
	&D_{part}\left(D_{ilbs}\left(\mathbf{x}, \beta, \theta, \psi\right), \theta, \psi\right)
\end{aligned}
\end{equation}
where $D_{ilbs}$ is the inverse linear blend skinning (LBS) operation based on the deformable FLAME head model~\cite{li2017learning},
and $D_{part}$ is the part-based deformation field to account for the fine-level motions, which will be introduced in the following.

\subsection{Skinning-based Coarse Deformation}

\label{subsec:coarse}

\paragraph{FLAME Model}

We adopt the parametric FLAME \cite{li2017learning} model as the geometry prior to perform skinning-based deformation.
FLAME consists of $N=5023$ vertices, $K=4$ joints (neck, jaw, and eyeballs) and $3$ types of blendshapes(including shape, pose dependent corrective and expression blendshapes).
The model can be defined as:
\begin{equation}
\begin{aligned}
	Flame(\beta, \theta, \psi) &=
	W\left(T_{P}(\beta, \theta, \psi), \mathbf{J}(\beta), \theta, \mathcal{W}\right) \\
	&=M(\mathbf{J}(\beta), \theta, \mathcal{W}) \cdot T_{P}(\beta, \theta, \psi)
	\label{func:flame_model}
\end{aligned}
\end{equation}
where
\begin{equation}
\begin{aligned}
	T_{P}(\beta, \theta, \psi)&=\overline{\mathbf{T}}+T_{PB}(\beta, \theta, \psi)\\
	T_{PB}(\beta, \theta, \psi) &= B_{S}(\beta ; \mathcal{S})+B_{P}(\theta ; \mathcal{P})+B_{E}(\psi ; \mathcal{E})
	\label{func:flame_tp}
\end{aligned}
\end{equation}
$W$ is a standard blend skinning function that rotates the vertices in the template mesh $\overline{\mathbf{T}} \in \mathbb{R}^{3 N}$ around joints $\mathbf{J} \in \mathbb{R}^{3 K}$ with transformation matrix $M$ linearly smoothed by blendweights $\mathcal{W} \in \mathbb{R}^{k \times N}$.
Parameters $\beta$, $\theta$, and $\psi$ are shape, pose, and expression coefficients. 
$B_{S}(\beta ; \mathcal{S}): \mathbb{R}|\beta| \rightarrow \mathbb{R}^{3 N}$ is the shape blendshapes to account for identity related shape variation.
$B_{P}(\theta ; \mathcal{P}): \mathbb{R}|\theta| \rightarrow \mathbb{R}^{3 N}$ is the corrective pose blendshapes to correct pose deformations that cannot be explained solely by linear bland skinning (LBS). 
$B_{E}(\psi ; \mathcal{E}): \mathbb{R}|\psi| \rightarrow \mathbb{R}^{3 N}$ is the expression blendshapes to capture facial expressions. 

\paragraph{Inverse LBS}

In FLAME, LBS is applied to deform a vertex position from the template mesh to animated mesh. 
In contrast, we build an inverse LBS model as our coarse deformation field to deform a spatial position from the posed space (video frame) to the canonical space (template fields). 
The template pose and expression in the canonical space is the “zero pose” $ \theta ^{\ast } $ of FLAME and the template shape is the average shape of all the frames computed by EMOCA~\cite{danecek2022emoca}. 
As there are only $5023$ surface vertices in FLAME, the dense points in the posed space are deformed following the movements of their k-nearest-neighbor vertices. Different from the deformation in \cite{chen2021animatable}, not only joints rotation but also the expression transformation and joints correction are important in Eq (\ref{func:flame_model}, \ref{func:flame_tp}). Thus $D_{ilbs}$ need to inverse both rotation transformations:

\begin{equation}
\begin{aligned}
	D_{ilbs}\left(\mathbf{x}, \beta, \theta, \psi\right)&=M^{-1}\left(\mathbf{x}, \beta, \theta\right) \cdot \mathbf{x} + T_{P}^{-1}(\beta, \theta, \psi) \\
	M^{-1}\left(\mathbf{x}, \beta, \theta\right) &=\mathop{AvgW}\limits_{v_{i} \in \mathcal{N}^{k}(\mathbf{x})}\left(M\left(\mathbf{J}(\beta), \theta, \mathcal{W}\right)[i]\right)^{-1}\\
	T_{P}^{-1}(\beta, \theta, \psi) &=\mathop{AvgW}\limits_{v_{i} \in \mathcal{N}^{k}(\mathbf{x})}(-T_{PB}(\beta, \theta, \psi)[i])
	\label{func:inverse_lbs}
\end{aligned}
\end{equation}

where $\mathcal{N}^{k}(\mathbf{x})$ is the set of $k$ nearest FLAME vertices of $\mathbf{x}$ in the posed space. $\mathop{AvgW}\limits_{v_{i} \in \mathcal{N}^{k}(\mathbf{x})}$ is the same weighted average function in \cite{chen2021animatable} and $[i]$ means the rotation matrix or transformation of $i$th vertex.
$M^{-1}$ is the inverse rotation matrix from Eq (\ref{func:flame_model}), and $T_{P}^{-1}$ is the inverse position transformation from Eq (\ref{func:flame_tp}) with 
the average shape $\beta_{avg}$, observed pose $\theta_{o}$ and expression $\psi_{o}$.

\subsection{Part-based Deformation Field}
\label{subsec:part-aware}

\paragraph{Motivation} Although the skinning-based deformation enables the point transformation from the posed space to the canonical space, it has the following deficiencies that limit its capacity for motion modeling.
\textit{First}, the predicted FLAME parameters inevitably contain estimation errors, making the skinning-based deformation prone to misalignment. 
\textit{Second}, the LBS weights used to compute the transformation are only defined on the surface, where some loose parts (\eg, hairs and glasses) are not well-modeled. 
\textit{Third}, the skeleton-based inverse LBS is incapable of dealing with subtle facial motions and fine-scale geometry details. 

A possible solution to address these shortcomings is to employ a global deformation network to predict a per-point offset to correct the coarsely-transformed point. 
However, as the facial motions and geometric details are inherently dissimilar between different facial parts, simply utilizing a global network is incapable to learn the distinctive part-based deformations and tends to produce an average result. 
This motivates us to propose a part-based deformation field to capture the fine-scale motion details.

\paragraph{Adaptive Part Assignment}
To successfully model the local motions of different face regions, we need to find a good partition for the space that accounts for the characteristics of the face geometry and motions.
A heuristic way is to manually partition the FLAME vertices into $N$ parts based on facial semantics~\cite{basso2008fitting,Kakadiaris2007ThreeDimensionalFR}, and assign points in the canonical space with the part labels of the nearest FLAME vertices. 
However, this hard part assignment strategy is sub-optimal as it is sensitive to the initial heuristic vertices partitions, and the solid boundaries between different parts may result in discontinuities in the deformation field.

To overcome the weaknesses of the above heuristic solution, we propose an adaptive part assignment mechanism, in which the points are adaptively assigned to the optimal parts along the training process.
In specific, we first build up the part-based deformation field with $N$ local deformation networks parameterized by $N$ tiny MLPs $\{\mathcal{D}_i, i={1..N}\}$.
For an input point $\mathbf{x}'$, the $i$-th MLP $D_i$ outputs a deformation offset for $\mathbf{x}'$ conditioned on the pose and expression parameter $(\theta, \psi)$, as the facial motions are usually pose- and expression-dependent.
We then adopt a part assigner MLP $\mathcal{F}_{assigner}$ to predict semantic logits $S = \mathcal{F}_{assigner} \left( x', \theta, \psi \right) \in \mathbb{R}^N$, which indicates the probabilities of the point belonging to different parts.
Therefore, the part-based deformation for point $\mathbf{x}'$ is defined as:
\begin{equation}
	D_{part}\left( \mathbf{x}', \theta, \psi \right) = \sum_{i=1}^{N} \mathcal{D}_i\left( \mathbf{x}', \theta, \psi \right) \times S_i,
\end{equation}
where $S_i$ refers to the $i$-th element of the predicted probability, and $\mathbf{x}' = D_{ilbs}\left(\mathbf{x}, \beta, \theta, \psi\right)$ is the previous coarsely-deformed point. 
\begin{figure*}[!t] \centering
    \makebox[0.15\textwidth]{\small FOMM}
    \makebox[0.15\textwidth]{\small SAFA}
    \makebox[0.15\textwidth]{\small NerFace}
    \makebox[0.15\textwidth]{\small IMavatar}
    \makebox[0.15\textwidth]{\small Ours}
    \makebox[0.15\textwidth]{\small GT}
    \\
    \includegraphics[width=0.9\textwidth]{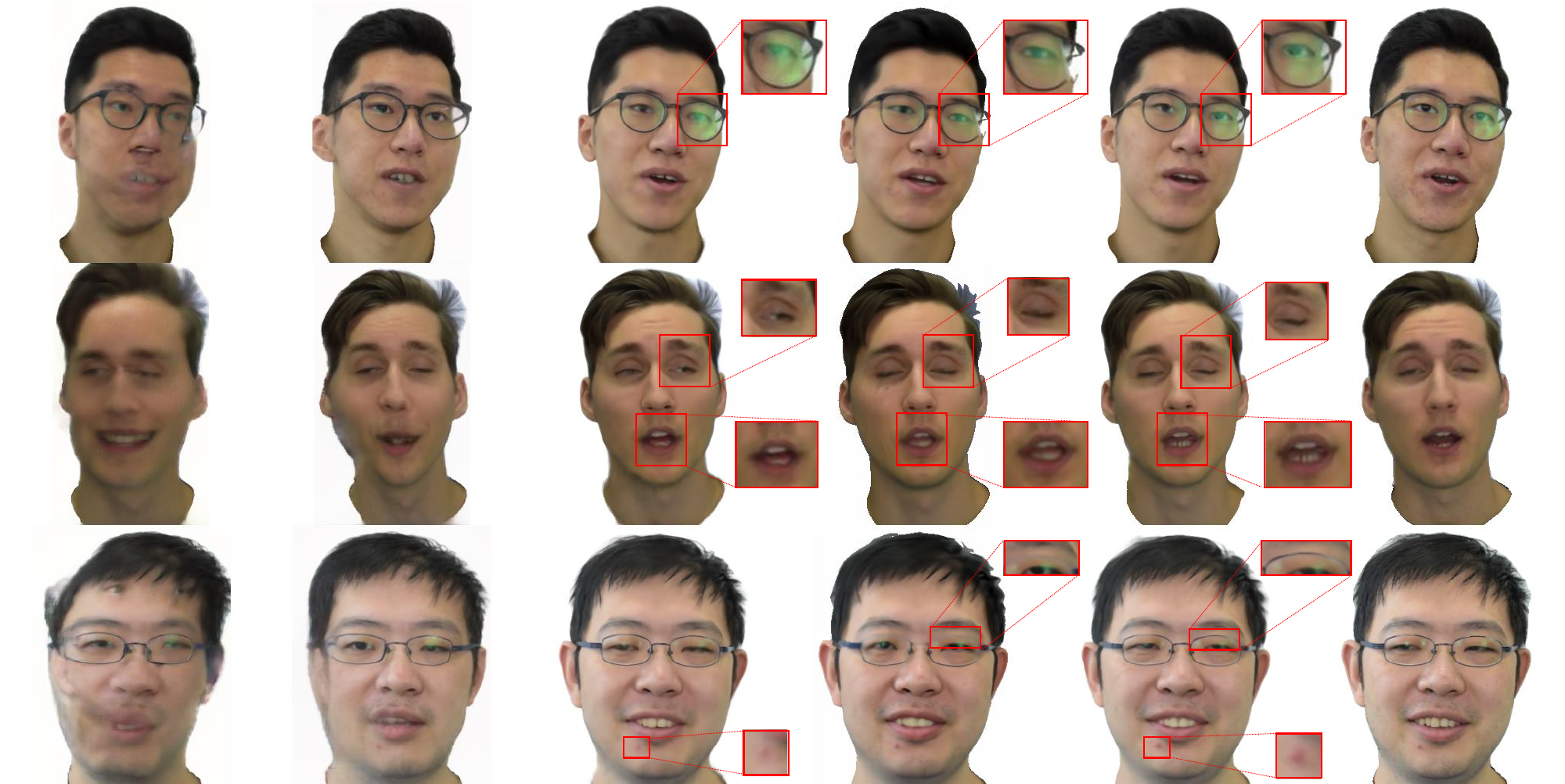}
    \caption{\textbf{Qualitative comparison on the reconstruction of Testset}. Some regions are zoomed in for detailed comparisons.}
\label{fig:comp_testset}
\vspace{-5mm}
\end{figure*}

\paragraph{Training Strategy}
Since the adaptive part assigner is trained fully unsupervised, we experimentally find that directly optimizing the part assigner MLP, local deformation networks, and the canonical radiance field from scratch will cause capacity degradation in local deformation networks, \ie, the space partition produced by the part assigner is less meaningful, and the deformation field will be dominated by some of the local networks. 
To solve this problem, we adopt a two-stage optimization strategy. 
we first utilize a heuristic hard part assignment strategy (based on the semantic labels of the vertices in the FLAME model) to obtain an initial partition for each point, which serves as a pseudo label, then solely optimize the local deformation networks and canonical radiance fields.
Despite the discontinuity of hard part assignment, this warm-up training provides an adequate initialization for local deformation networks.
Then, we discard the pseudo label while bring back the part assigner and jointly optimize all components. 
Our two-staged optimization strategy avoids the degradation of the part assigner and local deformation networks, and produces meaningful space partitions (see Ablation Study). 


\subsection{Volume Rendering and Loss Function}
We render the canonical radiance fields using the re-formulated volume rendering equation described in~\cite{oechsle2021unisurf}.
Given a ray $\mathbf{r} = \mathbf{o} + t\mathbf{d}$ in the posed space ($\mathbf{o}$ denotes the camera location and $\mathbf{d}$ the view direction), we sample $K$ points $\{\mathbf{x}_i|i=1,\dots,K\}$ and map it to the canonical space as $\{\mathbf{x}^c_i|i=1,\dots,K\}$ with the proposed coarse-to-fine deformation.
Then the expected color for this ray is approximated by the numerical quadrature:
\begin{equation}
	\hat{C}(\mathbf{r})=\sum_{i=1}^{K}o\left(\mathbf{x}^c_{i}\right)\prod_{j=1}^{i-1}\left(1-o\left(\mathbf{x}^c_{j}\right)\right) \mathbf{c}\left(\mathbf{x}^c_{i}, \mathbf{n}^c_i, \mathbf{d} \right)
\end{equation}
where $\mathbf{n}^c_i$ is the normal at $\mathbf{x}^c_{i}$ and is given by $\mathbf{n}\left(\mathbf{x}\right)=\nabla_{\mathbf{x}}o\left(\mathbf{x}\right)/\left\|\nabla_{\mathbf{x}}o\left(\mathbf{x}\right)\right\|_{2}$ which can be derived from double backpropagation.

We employ the same loss function as in~\cite{oechsle2021unisurf}, which comprises a L1 photometric loss and a surface regularization term:
\begin{equation}
	\begin{aligned}
		\mathcal{L} = \sum_{\mathbf{r} \in \mathcal{R}}\left\|\hat{C}(\mathbf{r})-C(\mathbf{r})\right\|_{1} + \lambda * \sum_{\mathbf{x}_{s} \in \mathcal{S}}\left\|\mathbf{n}\left(\mathbf{x}_{s}\right)-\mathbf{n}\left(\mathbf{x}_{s}+\boldsymbol{\epsilon}\right)\right\|_{2}
	\end{aligned}
\end{equation}
in which $\mathcal{R}$ is the set of training rays, $\mathcal{S}$ is the set of surface points, $C(\mathbf{r})$ is the ground-truth color, $\lambda$ is a balancing weight, and $\epsilon$ is a small uniformly sampled 3D perturbation.

\section{Experiments}
\label{sec:exp}
In this section, we present experimental results to verify the effectiveness of the proposed method.



\begin{figure*}[!t] \centering
	\includegraphics[width=1.0\textwidth]{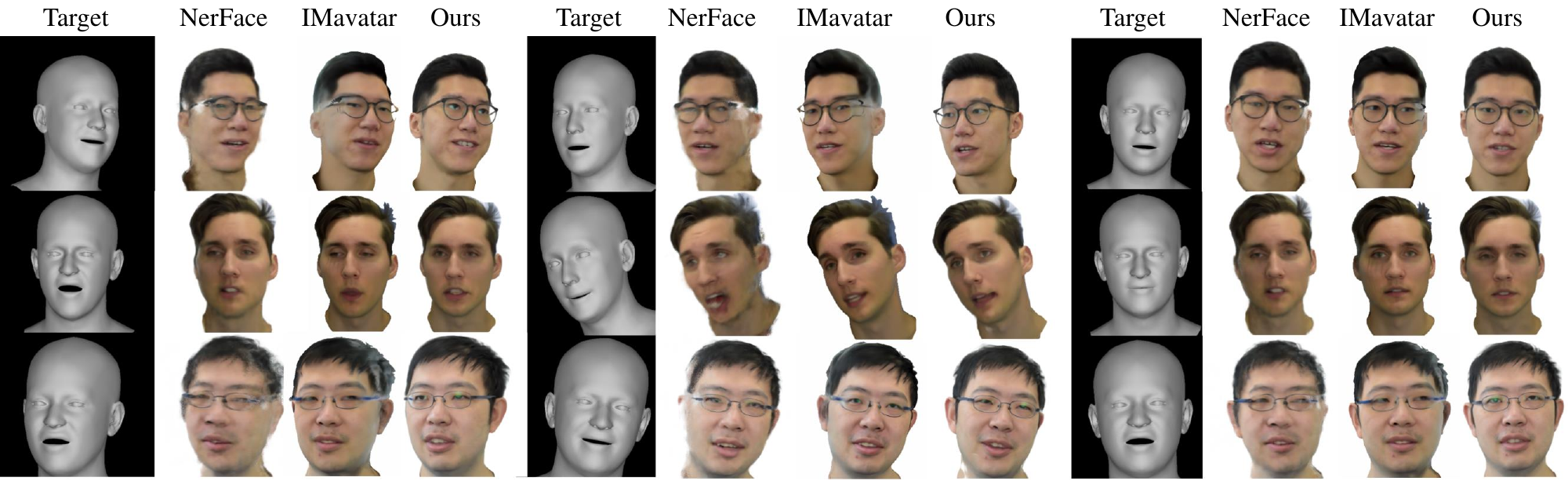}
	\caption{\textbf{Visual comparison of animation results with unseen poses and expressions.}} \label{fig:comp_unseen}
	\vspace{-5mm}
\end{figure*}

\paragraph{Implementation Details}
The canonical radiance fields use the same networks as in~\cite{oechsle2021unisurf}, which consists of an 8-layer MLP for predicting the occupancy $o$, and a 4-layer MLP to predict the color. The network structures for local deformation networks and part assigner is similar to the original NeRF~\cite{mildenhall2020_nerf_eccv20}, except we shorten the networks and add extra input as conditions. More details of network architectures are referred to the Sup. Mat. We adopt Adam~\cite{kingma2015adam} optimizer for the training. The learning rate starts from $5e^{-4}$ and decays exponentially to $5e^{-5}$ along the training process. We use a batch size of $4$ with $1024$ rays per-batch. The rays are unevenly sampled from foreground and background with a ratio of $9:1$. We set the hyper-parameter $|\mathcal{N}^{k}(\mathbf{x})|=4$ which is the number of nearest FLAME vertices in Eq.\ref{func:inverse_lbs} and $\lambda=0.005$. The training consumes about 24 hours on 2 Nvidia GeForce RTX 3090 24GB GPUs.

\paragraph{Datasets}
We evaluate our method on the public real face video dataset from~\cite{gafni2021dynamic}. The video dataset consists of 3 human subjects, captured from a single stationary camera at a resolution of $1920 \times 1080$ pixels. The images are cropped and resized to $512 \times 512$.
The videos include subjects engaging normal conversations, such as smiling and head rotations.
For each video, we use the last $1000$ frames for testing, while the remaining for training (roughly $5000$ frames), following the practice of~\cite{gafni2021dynamic}. 
We employ a segmentation network~\cite{ke2020modnet} to obtain the human masks and set the background to white color. 
The FLAME parameters are initially estimated using EMOCA~\cite{danecek2022emoca}, then we fix the expression and refine the rest parameters with 2D face landmarks loss. We kindly refer the reader to Sup. Mat. for more details.

\paragraph{Evaluation Metrics}
To compare the results of our reconstructed facial avatars with competitors, we evaluate the image quality of rendering results for all methods. Following the community standards~\cite{siarohin2019animating, wang2021safa, gafni2021dynamic}, the following metrics are measured: (1) \textit{L1 distance}, which calculates the average L1 distance between rendered images and ground truth images. (2) \textit{Peak Signal-to-noise Ratio (PSNR)}, which measures the image quality between rendered images and ground truths through mean squared error. (3) \textit{Structural Similarity Index (SSIM)}, which measures the structural similarity between two images. (4) \textit{Learned Perceptual Image Patch Similarity (LPIPS)}, which measures the perceptual similarity that has been shown to better match human perceptions~\cite{zhang2018unreasonable}.

\subsection{Comparison with State-of-the-arts}

\begin{table}[!t]
	\caption{\textbf{Quantitative Comparsion on the reconstruction of Testset}. We report mean values for each metrics. Although the quantitative results of NerFace are slightly better than ours, we argue this is due to its overfitting on training data. However, when feed with unseen pose and expressions, its quantitative results declined by a large margin (see supplemental materials).}
	\label{tab:comp_testset}
	\centering

	\begin{tabular}{ccccc}
	\toprule
	& PSNR $\uparrow$  & SSIM $\uparrow$ & L1 $\downarrow$   & LPIPS $\downarrow$ \\ 
	\midrule
	FOMM    & 22.22 & 0.87  & 0.085 & 0.035  \\
	SAFA    & 18.47 & 0.85  & 0.142 & 0.050  \\
	NerFace & \textbf{26.77} & \textbf{0.93} & \textbf{0.051} & \textbf{0.012}  \\
	IMavatar & 23.78  & 0.91   & 0.073  & 0.018 \\
	Ours    & 25.73 & \textbf{0.93}  & 0.056 & \textbf{0.012}  \\ 
	\bottomrule
	\end{tabular}
	\vspace{-3mm}
\end{table}

\paragraph{Baselines} 
We compare our method with several recent state-of-the-art methods: (1) \textit{FOMM}~\cite{siarohin2019animating}, which uses pure 2D motion to animate the facial avatar. (2) \textit{SAFA}~\cite{wang2021safa}, which employs a mixture of FLAME model and 2D motions. (3) \textit{NerFace}~\cite{gafni2021dynamic}, which use dynamic neural radiance fields for facial avatar rendering. (4)IMavatar~\cite{zheng2021avatar}, which learns a canonical radiance fields along with neural blend skinning for animation.
In all experiments, the one-shot methods (\ie, FOMM, SAFA) use the first frame of testset as source image, while the rest frames as drivings. We first conduct comparison on the testset of each subjects for all baselines. Then, we compare the results of extrapolation of unseen pose and expression for NerFace, IMavatar and ours. 

\paragraph{Results on Testset}
The qualitative and quantitative results are shown in~\Fref{fig:comp_testset} and \Tref{tab:comp_testset}, respectively. The synthesized results of image-based methods (FOMM, SAFA) suffer from huge distortions, as their 2D motion-based image warping is incapable of modeling head rotations and expressional motions. Both NerFace, IMavatar and our method generate photo-realistic images with correct head rotations and expressions. Furthermore, comparing to IMavatar, our method can reconstruct fine-scale details (e.g. glasses) beyond FLAME's naked head model, thanks to the proposed part-based deformations. 
Although the visual results of NerFace are slightly better than IMavatar and ours, we argue that their high-fidelity rendering results are mainly attributed to the overfitting of training pose and expressions, while admit a degenerate solution that perfectly explains the training images, but generalizes poorly to novel test pose and expressions~\cite{Zhang20arxiv_nerf++}. 
And the subsequent extrapolation experiment further demonstrate this.

\begin{figure*}[!t] \centering
    \makebox[0.11\textwidth]{Global}
    \makebox[0.11\textwidth]{Hard}
    \makebox[0.11\textwidth]{Soft}
    \makebox[0.11\textwidth]{Ours}
    \makebox[0.11\textwidth]{GT}
    \makebox[0.11\textwidth]{Hard}
    \makebox[0.11\textwidth]{Soft}
    \makebox[0.11\textwidth]{Ours}
    \\
    \includegraphics[width=.9\textwidth]{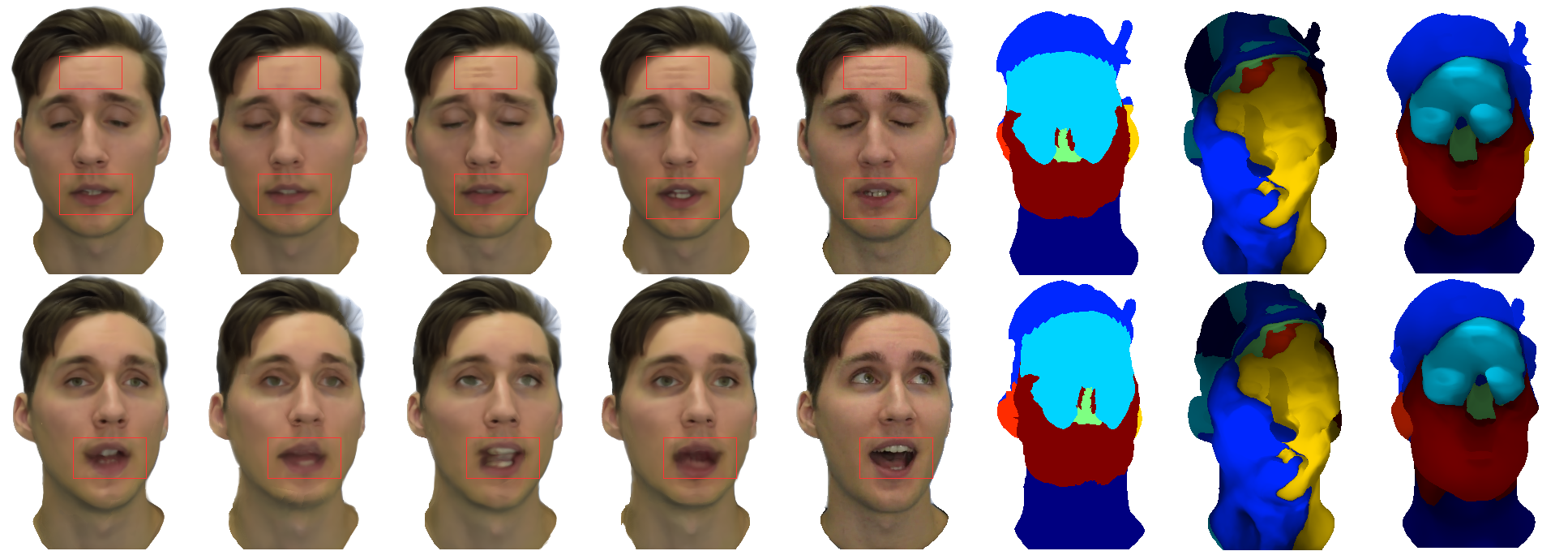}
\caption{\textbf{Qualitative comparison of ablation studies}. Columns 1-5: the visual comparison of between different variants and ground truths. Columns 6-8: the visualized face partitions on the learned surface. We use separate colors for each part and use the brightness to indicate the probability of surface points belonging this part (for hard part assignment, its probability is all ones).} 
\label{fig:ablate}
\vspace{-5mm}
\end{figure*}

\paragraph{Generalization to Unseen Pose and Expression}

To further investigate the reconstruction quality and generalization capability of our method. We randomly sample a set of unseen testing poses and expressions in their normal range of the corresponding continuous space, by making sure they are not existed in the training set (nor too close). We then feed this unseen testset into NerFace, IMavatar and ours. The qualitative comparison is shown in \fref{fig:comp_unseen}. Evidently, when extrapolating NerFace with out-of-domain pose and expressions, the performance of NerFace degrades drastically, indicating high risk of over-fitting. In contrast, IMavatar and our method present better generalization and robustness for unseen pose and expressions. We argue this superiority is likely attributed to the increased modeling capacity by our local part-based representations. Please refer to Sup. Mat. for more numerical analysis.

\subsection{Ablation Study}
\label{subsec:ablate}
To validate the main contributions of our proposed method, in this section, we conduct several ablation experiments for our method design. In particular, we discuss several alternative variants to our method, and compare them both quantitatively and qualitatively.

\begin{table}[!t]
	\caption{\textbf{Quantitative comparsion of ablation studies}. }
	\label{tab:ablation}
	\centering
	\begin{tabular}{ccccc}
		\toprule
		& PSNR $\uparrow$  & SSIM $\uparrow$ & L1 $\downarrow$   & LPIPS $\downarrow$ \\ 
		\midrule
		Global    & 25.30 & 0.92  & 0.059 & 0.014  \\
		Hard    & 25.60 & 0.92  & 0.061 & 0.013  \\
		Soft & 25.49 & 0.92 & 0.057 & 0.013  \\
		Ours    & \textbf{25.73} & \textbf{0.93}  & \textbf{0.056} & \textbf{0.012}  \\ 
		\bottomrule
	\end{tabular}
\vspace{-3mm}
\end{table}

\paragraph{Global deformation field}
Corresponding to our part-based deformation field, a straightforward way to model fine-level deformation is to learn a global deformation network that outputs offsets for all spatial points, as done in many body-related works~\cite{weng2022humannerf,zheng2022structured}. However, simply using a global function does not take into account the part-related face characteristics. To verify this, we design a variant of our method that employs a single deformation network to predict an offset for canonical points transformed by the skinning-based deformation. The input and the rest components are kept unchanged to fix the variables. 
Moreover, to eliminate the effect of limited network capacity, we enlarge the parameter number of the single deformation network to be $N$ times as our local deformation networks, reaching an equal number of network parameters.
\fref{fig:ablate} presents the visual comparison between this variant and our method. It is shown that the global deformation field is unable to capture fine-scale details and large facial motions and tends to produce blurry results.

\paragraph{Training strategies of part-based deformation fields}

There are two alternatives to the proposed two-stage training strategies: (1) \textit{pure hard part assignment, abbr. Hard}, in which the part assigner is removed, and each point's part ascription is obtained by querying the part label from the nearest vertex on FLAME model.
(2) \textit{jointly train part assigner from scratch, abbr. Joint}, in which all the components including part assigner, local deformation networks, radiance fields are jointly trained from scratch. We conduct experiments on these strategies with the same experimental setting as our main method. The visual comparisons are presented in~\fref{fig:ablate} (col. 2\&3). The rendering results for \textit{Hard} are blurrier and show discontinuity at the face-neck boundaries, while the \textit{Joint} variant exhibits some distortions under big expressions, which is likely to be caused by its incorrect face partitions. Furthermore, from the visualization of the learned face partitions (col. 6-8), our staged training strategy outputs physically plausible partitions that conforms with the facial dynamics, e.g. the hair part is accurately aligned and the merged forehead and eye regions indicating they are likely to have similar facial muscle dynamics.





\section{Conclusion}
In this paper, we have introduced the part-based neural radiance field (PartNerFace) for animatable facial avatar reconstruction from monocular RGB videos.
Our method first maps the observed points to the canonical space by inverse skinning based on the FLAME model, and then model fine-scale facial motion with a part-based deformation field.
Extensive experiments show that our method generalizes well to unseen expressions and is capable of rendering realistic facial details, clearly outperforming existing methods. 
\clearpage

\bibliography{reference}

\end{document}